%% file: main.tex
\documentclass[letterpaper, 10pt, conference]{ieeeconf}
\IEEEoverridecommandlockouts
\usepackage{amsmath}
\usepackage{amsfonts}
\usepackage{amssymb}
\usepackage{bm}
\usepackage{graphicx}
\usepackage{siunitx}
\usepackage{booktabs}
\usepackage[table]{xcolor}
\usepackage{multirow}
\usepackage{algorithm}
\usepackage{algpseudocode}
\usepackage{cite}
\usepackage{url}
\usepackage{hyperref}
\hypersetup{hidelinks}

\makeatletter
\renewcommand\fs@ruled{\def\@fs@cfont{\bfseries}\let\@fs@capt\floatc@ruled
  \def\@fs@pre{\kern10pt\hrule height.8pt depth0pt \kern2pt}%
  \def\@fs@post{\kern2pt\hrule\relax}%
  \def\@fs@mid{\kern2pt\hrule\kern2pt}%
  \let\@fs@iftopcapt\iftrue}
\makeatother
\floatstyle{ruled}
\restylefloat{algorithm}

\title{\LARGE \bf Terrain-Grounded Imagination: A Recipe for\\ Training Perceptive Legged Locomotion through a World Model}

\author{Matthias Heyrman$^{*}$ and Marco Hutter\\[2pt]
{\normalsize Robotic Systems Lab, ETH Z\"urich, Switzerland}%
\thanks{$^{*}$Corresponding author: {\tt\footnotesize mheyrman@ethz.ch}}%
\thanks{Project page: {\protect\urlstyle{same}\protect\href{https://github.com/leggedrobotics/terrain-grounded-imagination}{\protect\nolinkurl{github.com/leggedrobotics/terrain-grounded-imagination}}}}%
}

\makeatletter
\long\def\@makecaption#1#2{%
\ifx\@captype\@IEEEtablestring%
{\footnotesize\centering #1\par\vskip 1pt #2\par}%
\@IEEEtablecaptionsepspace%
\else
\@IEEEfigurecaptionsepspace%
\setbox\@tempboxa\hbox{\footnotesize #1.~~ #2}%
\ifdim \wd\@tempboxa >\hsize%
\setbox\@tempboxa\hbox{\footnotesize #1.~~ }%
\parbox[t]{\hsize}{\footnotesize \noindent\unhbox\@tempboxa#2}%
\else%
\ifcenterfigcaptions \hbox to\hsize{\footnotesize\hfil\box\@tempboxa\hfil}%
\else \hbox to\hsize{\footnotesize\box\@tempboxa\hfil}%
\fi\fi\fi}
\makeatother

\begin{document}

\bstctlcite{IEEEexample:BSTcontrol}

\maketitle
 \thispagestyle{empty}
\pagestyle{empty}

\input{sections/s0_abstract}
\input{sections/s1_introduction}

\input{sections/s2_related}
\input{sections/s3_method}
\input{sections/s4_experiments}
\input{sections/s5_conclusion}

\section*{ACKNOWLEDGMENT}
The authors would like to thank the members of the Robotic Systems Lab for their advice during this project, and Chenhao Li for his insights into world modeling for robotics.
The authors acknowledge the use of generative AI to assist with code generation and language refinement.

\bibliographystyle{IEEEtran}
\bibliography{references}

\end{document}

%% file: sections/s0_abstract.tex
\begin{abstract}
Model-free reinforcement learning for perceptive legged locomotion queries the simulator up to billions of times to calculate state transitions at every time step.
A world model is a learned, differentiable transition function that requires fewer simulator steps to learn, and model-based reinforcement learning uses it to reduce collection time.
Typically, such a model would also have to predict the environment that the robot walks on and the resulting perceptive inputs for the policy.
However, the terrain is generated before training and is therefore already known in simulation.
We present terrain-grounded imagination (TGI), in which the world model only learns how the robot responds to a pre-generated terrain, and rewards, terminations, and robot perception are computed from differentiable queries of the terrain geometry.
New terrain or perception changes the query rather than what the world model learns.
Policies are trained on 3D terrains that require contact-rich locomotion for climbing, and we demonstrate using 3D perception to pass under overhangs.
A world model that predicts the contact force on every body of the robot also makes rewards on contact forces available in imagination, and cutting the gradient at terrain edges reduces falls.
Compared to training in simulation with proximal policy optimization (PPO), TGI policies match or exceed its success rate, with more falls for the humanoid.
Training for the same number of iterations takes $6.1\times$ less time for humanoid locomotion on 2.5D terrain and $4.9\times$ less time for contact-rich quadruped locomotion, for which PPO uses default settings.
\end{abstract}

%% file: sections/s1_introduction.tex
\section{Introduction}

\suppressfloats[t]
\begin{figure}[t]
    \centering
    \includegraphics[width=\linewidth]{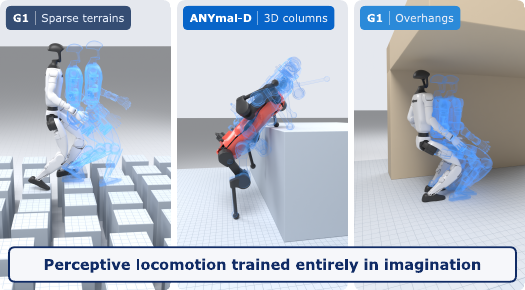}
    \caption{TGI trains perceptive locomotion in imagined rollouts (blue) of a learned world model, with the terrain queried from the known geometry: a height map for the G1 on sparse terrain, columns for ANYmal-D on 3D terrain, and voxels for the G1 under overhangs.}
    \label{fig:teaser}
\end{figure}

Reinforcement learning (RL) in simulation has become the standard approach for training legged locomotion controllers~\cite{pmlr-v164-rudin22a, doi:10.1126/scirobotics.abk2822}.
Model-free algorithms such as proximal policy optimization (PPO)~\cite{PPO} treat the simulator as a black box and query it up to billions of times.
Learned world models reduce these queries.
SuperMimic~\cite{liao2026supermimic} fits a world model to simulator transitions and backpropagates the return through short imagined rollouts.
It trains a general humanoid motion tracker with three orders of magnitude fewer GPU-hours than the PPO-based SONIC~\cite{sonic}.
These results, however, were obtained on flat ground, and most other world models for locomotion have been used to train blind policies~\cite{li2025rwm} or as representations for policies trained with PPO in a simulator~\cite{lai2025wmp, liu2026wmloco}.

Training a perceptive policy entirely through a world model raises two challenges.
First, imagined rollouts must reproduce the terrain for locomotion and the robot's view of it to serve as an input to the policy, which the world model would have to learn in addition to the dynamics of the robot.
Second, sparse terrain such as stepping stones produces many contact discontinuities: when a foot lands on an edge, a small change in the action decides between a step and a fall, and first-order gradients through such dynamics have high variance and bias~\cite{suh2022differentiable,schwarke2025learning,zhang2023model}.

We address the first challenge by grounding imagination in the terrain, which is known in simulation.
In terrain-grounded imagination (TGI, Fig.~\ref{fig:teaser}), the world model learns only the robot's response to the terrain, which it reads through a differentiable query of the simulator's geometry at points on the robot.
Forward models for navigation are likewise conditioned on local elevation maps~\cite{roth2025learned}.

Since the terrain enters only through the query, a new kind of terrain or perception changes the query rather than what the world model learns (Fig.~\ref{fig:overview}).
For a humanoid on stairs, gaps, and stepping stones, it is a height lookup under the base and the soles.
For a quadruped that climbs obstacles of its own height, each node of the height map becomes a column, which gives every body its distance to the nearest surface and the surface normal, walls included.
From this contact geometry, an impulse world model learns the force on every body, so hard impacts and body contacts can be penalized in imagination.
For overhangs, a second layer represents the geometry that does not rest on the ground.

The known terrain also lets us address the second challenge.
Because the support of every imagined foothold can be queried, we can locate the terrain edges where the return changes abruptly with the action and its gradient becomes unreliable.
We therefore cut the gradient at these poorly supported touchdowns and let the critic estimate the return after them, which reduces the falls of the trained policy.

We demonstrate TGI on two robots: the Unitree G1 humanoid and the ANYmal-D quadruped.
The humanoid case shows that a training setup tuned for PPO carries over to world model training on rough and sparse terrain, such as stairs, gaps, and stepping stones. Training to \num{40000} iterations is $6.1\times$ faster than with PPO, and the policy is deployed on hardware.
For contact-rich quadruped locomotion, where the legs and the base also touch the terrain, ANYmal-D is trained to navigate over climbs, gaps, stairs, and stepping stones. Training to \num{30000} iterations is $4.9\times$ faster than with PPO at default settings, and TGI reaches the final curriculum level of PPO with $10\times$ fewer simulator transitions.

In summary, our contributions are:
\begin{enumerate}
    \item Terrain-grounded imagination (TGI), a method for training perceptive legged locomotion entirely in a world model that learns only the robot, while its contacts, perception, and rewards are computed from differentiable queries of the known terrain (Sec.~\ref{sec:method}).
    \item Terrain queries for 2.5D and 3D terrain that reproduce the robot's perception in imagination, with which humanoid and quadruped locomotion tasks carry over from simulation to world model training (Sec.~\ref{ssec:lookup}).
    \item Two task-specific additions: an impulse world model that predicts the contact force on every body of the robot, so that force penalties can be computed in imagination (Sec.~\ref{ssec:wm}), and a gradient cut at poorly supported touchdowns that reduces falls (Sec.~\ref{ssec:objective}).
\end{enumerate}

%% file: sections/s2_related.tex
\begin{figure*}[t]
    \centering
    \includegraphics[width=\linewidth]{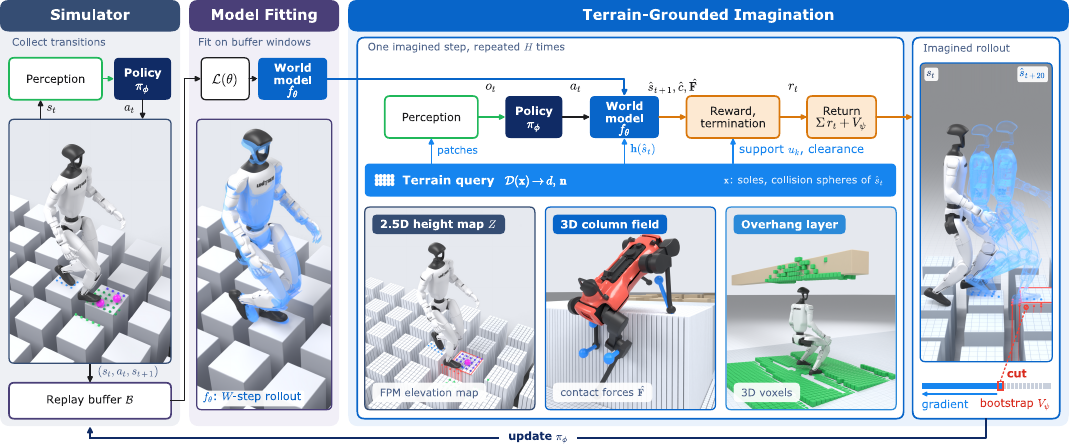}
    \caption{Overview of TGI. The policy $\pi_\phi$ collects simulator transitions in a replay buffer $\mathcal{B}$ (left), on which the world model $f_\theta$ is fit (middle). The policy is improved on imagined rollouts (right), in which the model predicts the robot and the terrain query $\mathcal{D}$ supplies the observation, the terrain features $\mathbf{h}$, the rewards, and the terminations, for 2.5D height maps, 3D column fields, and overhangs (bottom). The gradient is cut at poorly supported touchdowns, where the return is bootstrapped from the critic $V_\psi$.}
    \label{fig:overview}
    \vspace{-2pt}
\end{figure*}

\section{Related Work}
\label{sec:related}

Model-based RL learns a model of the environment that predicts state transitions that are used to train a policy.
The policy can learn from these predicted transitions~\cite{sutton1991dyna}, from rollouts branched from real recorded states~\cite{janner2019mbpo}, or entirely in the imagination of the model~\cite{hafner2025dreamerv3}, as DayDreamer did for a quadruped that learned to walk directly on hardware~\cite{wu2022daydreamer}.
For legged robots, world models have mostly been used as learned simulators of the robot's dynamics.

The Robotic World Model (RWM) trained proprioceptive velocity-tracking policies with PPO on long autoregressive rollouts~\cite{li2025rwm} or from offline data~\cite{li2026rwmu}.
Because a learned model is differentiable, the return of an imagined rollout can be backpropagated to the policy.
SuperTrack trained physics-based character controllers in this way~\cite{fussell2021supertrack}, and SuperMimic extended this to train a general humanoid motion tracker deployed on a Unitree G1~\cite{liao2026supermimic}.
All of these have trained blind policies that do not need the world model to predict the policy's visual inputs.

Perceptive legged locomotion is commonly learned with model-free RL in massively parallel simulation~\cite{pmlr-v164-rudin22a} from elevation maps~\cite{doi:10.1126/scirobotics.abk2822}.
With these, humanoids have traversed sparse footholds, platforms, and gaps~\cite{wang2025beamdojo,zhuang2024humanoid} and quadrupeds have climbed and jumped over large obstacles~\cite{hoeller2024anymalparkour}, and 3D representations extend this to terrain with overhangs~\cite{miki2024confined,ben2025gallant}.
Off-policy learning at scale is one route to faster training~\cite{seo2025learning,kim2026flashsac}.
For perceptive locomotion, world models have mostly served as representations for policies that are still trained in simulation~\cite{lai2025wmp,liu2026wmloco} instead of replacing the simulator.

Recent work brings the terrain into a world model in which the policy is trained.
DynaMimic gives local height maps to the world model of a humanoid that tracks reference motions~\cite{xing2026dynamimic}.
TGI trains the policy entirely in imagination and without reference motions, and its exact terrain query supplies the policy's perception and the support of every foothold in addition to the dynamics, which extends training in a world model to stairs, gaps, and stepping stones, and to 3D terrain.

To make the learned dynamics more accurate, other models build rigid-body structure into the prediction.
Semi-structured models learn only the external torques within known Lagrangian dynamics~\cite{levy2024ssrl,huang2026lift}.
DynaMimic also predicts contact wrenches at candidate support links, which enter closed-form forward dynamics~\cite{xing2026dynamimic}.
NeRD replaces the dynamics and contact solvers of an analytical simulator with a learned model and keeps the simulator's collision detection~\cite{xu2025nerd}.
Our impulse model also maps predicted forces through the contact Jacobian, but takes the contact geometry from the differentiable terrain query, so imagination needs no simulator.

Backpropagating the return gives the policy a first-order gradient, which is reliable only where the dynamics are smooth~\cite{song2024learning}.
Through contacts, these gradients suffer from high variance and bias~\cite{suh2022differentiable,parmas2018pipps}.
SHAC backpropagates only through short windows of a differentiable simulator and bootstraps the remainder with a learned critic~\cite{xu2022shac}, and AHAC adapts the window to avoid stiff contacts~\cite{georgiev2024ahac}.
PWM computes such first-order gradients through a learned world model instead of a simulator~\cite{georgiev2025pwm}.
The same problem appears in learned world models: for a humanoid motion tracker, the imagined rollouts with the largest gradients predicted changes in the return that the simulator did not reproduce~\cite{dang2026trust}.
Unlike AHAC, TGI cuts the gradient only where the terrain query detects a foot landing on an edge.

%% file: sections/s3_method.tex
\section{Terrain-Grounded Imagination}
\label{sec:method}

\subsection{Overview}
\label{ssec:setting}

We train a perceptive locomotion policy $\pi_\phi(a_t \mid o_t)$ in a world model that is learned from simulation.
The robot state $s = (\mathbf{p}, \mathbf{R}, \mathbf{v}, \bm{\omega}, \mathbf{q}, \dot{\mathbf{q}})$ consists of the pose and velocities of the base and the joint states.
The action $a$ sets joint targets $\mathbf{q}^{*}$ for proportional-derivative (PD) control at \SI{50}{\hertz}, and the physical parameters $\xi$ are randomized per environment.
Training follows other model-based methods~\cite{sutton1991dyna,janner2019mbpo,li2025rwm,liao2026supermimic}: the policy collects simulator transitions in a replay buffer $\mathcal{B}$, the world model is fit to $\mathcal{B}$, and the policy is trained on rollouts imagined from states in $\mathcal{B}$ (Fig.~\ref{fig:overview}).
In TGI, the world model only predicts the next state of the robot,
\begin{equation}
    \hat{s}_{t+1} = f_\theta\big(\hat{s}_t, a_t, \mathbf{h}(\hat{s}_t), \xi\big),
    \label{eq:model}
\end{equation}
where $\mathbf{h}$ are features of the terrain around the robot.
The model is deterministic and conditioned on $\xi$, as the sources of randomness in simulation are known~\cite{liao2026supermimic}.
Since the terrain is generated before training, we compute the terrain features, the rewards, the terminations, and the observation $o_t$ from a differentiable query of its geometry.

\subsection{Terrain Query}
\label{ssec:lookup}

The terrain query $\mathcal{D}(\mathbf{x})$ returns the clearance $d$ between a point $\mathbf{x} \in \mathbb{R}^3$ and the terrain, and the normal $\mathbf{n}$ of the nearest surface.
It is differentiable in $\mathbf{x}$, so the policy gradient can be backpropagated through it.

For 2.5D terrain such as stairs, gaps, and stepping stones, where only the feet touch the ground, we use the height below the point (2.5D TGI):
\begin{equation*}
    d = x^{z} - \mathcal{T}(\mathbf{x}^{xy}), \quad \mathbf{n} = \mathbf{e}_z,
\end{equation*}
where $\mathcal{T}$ bilinearly interpolates a height map $\mathbf{Z}$ of resolution $\delta$ that is sampled once from the terrain mesh.
A finer height map gives sharper edges but larger gradients.
To evaluate the support of a foot, the exact edge is needed, so the lookup returns the nearest node and keeps the gradient of the interpolation~\cite{bengio2013estimating}.

On terrain that the robot climbs, the legs and the base also touch vertical surfaces, which the height below a point does not describe.
We therefore treat each node of $\mathbf{Z}$ as a column that extends downward from its height, and $d$ and $\mathbf{n}$ are the signed distance to the nearest column and the normal of its surface (3D TGI), which gives every body of the robot its distance to walls as well as to the ground.

Columns cannot represent terrain with free space below it.
For overhangs, we add a second layer, a signed-distance grid with \SI{5}{\centi\metre} voxels, and the query returns the closer of the two layers.

\subsection{Terrain Features}
\label{ssec:features}

The terrain features $\mathbf{h}(s)$ describe the terrain around the robot to the world model.
They are terrain queries at points attached to the robot, which are positioned with differentiable forward kinematics.
In 2.5D TGI, they are the height of the base, the clearance of each foot, and the heights on a grid ahead of the base and across each sole.
3D TGI adds the clearance and the normal at each foot and at collision spheres on the thighs, knees, shanks, and base (Table~\ref{tab:params}).
Since the features are local to the robot, the world model does not depend on the layout or the size of the terrain.

\subsection{World Model}
\label{ssec:wm}

The world model $f_\theta$ is a multilayer perceptron (MLP) with the inputs $\big(\varsigma(s_t),\allowbreak \mathbf{h}(s_t),\allowbreak \mathbf{q}^{*}_t,\allowbreak \mathbf{q}^{*}_{t-1},\allowbreak \xi\big)$.
The robot state enters as the base tilt, base-frame velocities, and joint states $\varsigma(s)$, which makes the model invariant to position and heading.
The action $a_t$ enters as the joint targets $\mathbf{q}^{*}_t$, and the previous targets $\mathbf{q}^{*}_{t-1}$ account for the actuation delay.
We use two versions of the model.
The standard model predicts the state increment $\hat{s}_{t+1} \ominus \hat{s}_t$ in the base frame and the contact probability $\hat{c}_k$ of each foot $k$.

The impulse model predicts contact forces, using the contact geometry from the terrain query.
For each contact body $k$ with clearance $d_k$ and normal $\mathbf{n}_k$, the network outputs a logit $\ell_k$, a magnitude $m_k \geq 0$, and a tangential vector $\mathbf{t}_k \perp \mathbf{n}_k$, which define the contact force
\begin{equation}
\begin{aligned}
    \hat{c}_k &= \operatorname{sig}\big( (\epsilon_c - d_k) / \kappa_c \big) \operatorname{sig}(\ell_k), \\
    \hat{\mathbf{F}}_k &= \hat{c}_k\, m_k \big( \mathbf{n}_k + \mu \tanh(\|\mathbf{t}_k\|)\, \mathbf{t}_k / \|\mathbf{t}_k\| \big),
\end{aligned}
\label{eq:force}
\end{equation}
where $\operatorname{sig}$ is the logistic function.
A body away from the terrain has no force, a friction cone with a fixed $\mu = 1$ bounds the tangential force, and each force acts at the terrain-facing point of the body's nearest collision sphere.
The forces are mapped to the state increment through the contact Jacobian $\mathbf{J}_{\mathrm{c}}$,
\begin{equation*}
    \hat{s}_{t+1} \ominus \hat{s}_t = \bm{\Delta}^{\mathrm{f}} + \mathbf{B}\, \mathbf{J}_{\mathrm{c}}^\top \hat{\mathbf{F}},
\end{equation*}
where the network predicts the increment of the free motion $\bm{\Delta}^{\mathrm{f}}$ and the matrix $\mathbf{B}$, which maps the generalized contact force to the state increment.

Both models are trained on windows of $W$ transitions from the buffer.
The model is rolled out autoregressively with the recorded actions and with $\mathbf{h}$ recomputed from the predicted states, so that it learns from its own compounding errors~\cite{li2025rwm,liao2026supermimic}.
The loss is the normalized squared error of the state increments, the cross-entropy of the contacts with weight $w_c$, and, for the impulse model, the error of the forces in units of \SI{100}{\newton} with weight $w_f$.

\subsection{Imagined Rollout}
\label{ssec:perception}

Rewards and terminations are computed from the imagined state, the predicted contacts and forces, and the query, so no reward model is learned.
Terminations are products of sigmoids, which keeps them differentiable.
In 3D TGI, the clearances of the collision spheres are used to penalize contacts of the knees and shanks and to terminate the episode on a contact of the base.

The query also provides the support of each foot.
We measure it with the unsupported share,
\begin{equation*}
    u_k(s) = \frac{1}{N_{\mathrm{s}}} \sum_{m=1}^{N_{\mathrm{s}}} \operatorname{sig}\big( (|h_{k,m}| - \epsilon_u) / \kappa_u \big),
\end{equation*}
where $h_{k,m}$ are the $N_{\mathrm{s}}$ terrain heights across the sole relative to the height under the foot center, so that $u_k$ is the soft share of them that deviate by more than $\epsilon_u$.
A touchdown is poorly supported if $u_k > \eta$.

The observation of the policy is also computed from the query.
For perception, we adopt an existing method, the foothold-gated perception of~\cite{heyrman2026watch}.
A foothold prediction model (FPM) $g_\vartheta$ predicts the next foothold of each foot from the proprioceptive state, and the policy observes the noisy proprioceptive state, the previous action, the footholds, and a patch of $\mathbf{Z}$ around each foothold.
In imagination, the patches are read at the footholds predicted from the imagined state, which are treated as constants, so no observation model is learned.
For terrain with overhangs, the policy instead observes a voxel occupancy grid rendered by casting rays from the imagined sensor pose.

\begin{table}[t]
    \caption{Settings of the two TGI instances (LN: LayerNorm).}
    \label{tab:params}
    \centering\footnotesize
    \setlength{\tabcolsep}{3pt}
    \begin{tabular}{@{}l ll@{}}
        \toprule
        & G1 (2.5D TGI) & ANYmal-D (3D TGI) \\
        \midrule
        Task & velocity tracking & goal reaching \\
        Terrain query & height lookup & column field \\
        Base grid & $7 \times 5$ & $9 \times 5$ \\
        Foot grid & $5 \times 3$ & $3 \times 3$ \\
        Collision spheres & -- & 37 \\
        World model & standard & impulse \\
        Contact bodies & feet & feet, shanks, thighs, base \\
        Impact threshold & -- & \SI{600}{\newton} \\
        Force loss / $w_f$ & -- & squared error / 0.5 \\
        $\epsilon_c$ / $\kappa_c$ & -- & \SI{4}{\centi\metre} / \SI{1}{\centi\metre} \\
        Patch per foot & $7 \times 7$ & $5 \times 5$ \\
        Action scale & 0.25 & 0.5 \\
        Envs. / iterations & 8192 / \num{40000} & 4096 / \num{30000} \\
        \midrule
        \multicolumn{3}{@{}l}{\emph{Shared}} \\
        Height map resolution $\delta$ & \multicolumn{2}{l}{\SI{0.04}{\metre}} \\
        Support $\epsilon_u$ / $\kappa_u$ / $\eta$ & \multicolumn{2}{l}{\SI{3}{\centi\metre} / \SI{1}{\centi\metre} / 0.3} \\
        World model $f_\theta$ & \multicolumn{2}{l}{MLP $3 \times 1024$, LN, SiLU, Muon} \\
        $W$ / $w_c$ & \multicolumn{2}{l}{8 / 0.5} \\
        $\pi_\phi$ / $V_\psi$ (ELU) & \multicolumn{2}{l}{[512, 256, 128] / [512, 256]} \\
        Policy optimizer / $\beta$ & \multicolumn{2}{l}{Muon / 0.995} \\
        FPM $g_\vartheta$ & \multicolumn{2}{l}{[128, 128], LN, LeakyReLU} \\
        $H$ / rollouts / $\gamma$ / $\lambda$ & \multicolumn{2}{l}{32 / 2048 / 0.99 / 0.95} \\
        $w_{\mathrm{lr}}$ / action noise / $w_{\mathrm{m}}$ & \multicolumn{2}{l}{0.1 / 0.2 / 2.0} \\
        Learning rates & \multicolumn{2}{l}{$2 \cdot 10^{-4}$} \\
        Buffer / updates per step & \multicolumn{2}{l}{200 steps per env. / 0.4 per sim. step} \\
        \bottomrule
    \end{tabular}
\end{table}

\subsection{Policy Update}
\label{ssec:objective}

The policy maximizes the imagined return over a horizon $H$, with a critic $V_\psi$ trained on $\lambda$-returns~\cite{schulman2016gae,xu2022shac}.
The gradient of the return is backpropagated through the world model, the terrain query, and the rewards with reparameterized actions.
At a poorly supported touchdown, a small change in the contact position decides between a step and a fall, so the gradient is unreliable~\cite{suh2022differentiable}.

We detect these touchdowns with the terrain query as events $e_t$ and cut the gradient at them:
\begin{equation}
\begin{aligned}
    e_t &= \mathbf{1}\big[ \max_{k}\, \hat{c}_{k,t+1} (1 - \hat{c}_{k,t})\, u_k(\hat{s}_{t+1}) > \eta \big], \\
    G_t &= r_t + \gamma (1 - \rho_t) \big[ e_t V_\psi(\hat{s}_{t+1}) + (1 - e_t) G_{t+1} \big],
\end{aligned}
\label{eq:gated_return}
\end{equation}
where $\hat{c}_{k,t+1} (1 - \hat{c}_{k,t})$ marks a touchdown of foot $k$, $\rho_t$ is the termination probability, and $G_H = V_\psi(\hat{s}_H)$.
With the first event at $t^{*}$, $G_0$ is the return of~\cite{xu2022shac} with the horizon $\min(H, t^{*} + 1)$, so the cut gives a smoother objective at the cost of a bias toward the critic.
Actions after an event receive no backpropagated gradient, so we add a likelihood-ratio term on the normalized advantages of the $\lambda$-return without the cut, with weight $w_{\mathrm{lr}}$~\cite{parmas2018pipps,suh2022differentiable}, and a left-right mirror loss with weight $w_{\mathrm{m}}$~\cite{symmetry_aug,heyrman2026watch}.
The deployed policy is an exponential moving average of the trained policy with decay $\beta$.

\subsection{Instances and Training}
\label{ssec:training}

We apply TGI to two robots.
For humanoid locomotion on 2.5D terrain, a Unitree G1 tracks velocity commands in the environment of~\cite{heyrman2026watch}, which we port to Isaac Lab~3~\cite{nvidia2025isaaclabgpuacceleratedsimulation} with Newton physics~\cite{newton2025newton}.
We replace its stepping stones with strips of square stones as small as \SI{0.2}{\metre} with gaps of up to \SI{0.25}{\metre}.
For contact-rich quadruped locomotion, ANYmal-D uses 3D TGI in the rough-terrain environment of Isaac Lab with Newton.
It has to reach a goal \SIrange{2.5}{3.5}{\metre} away within \SI{8}{\second}~\cite{rudin2022advancedskillslearninglocomotion} with the reward of~\cite{zhang2026ame2agilegeneralizedlegged}.
The terrains are climbs of up to \SI{1.0}{\metre}, gaps, pits, boxes, stairs, and stepping stones.
Table~\ref{tab:params} lists the settings of both, and their randomization ranges are available in the code.
The G1 uses the standard world model, since its reward has no terms on contact forces, and ANYmal-D the impulse model with every contact body.

Both use success-guided sampling~\cite{zhang2026balanced} as the terrain curriculum.
The first 300 iterations train only the world model and the FPM, and until iteration 1500, the policy observes fixed footholds beside the base~\cite{heyrman2026watch}.
For the policy, the learning rate of Muon~\cite{jordan2024muon} decays from $4\times$ to $1\times$ the learning rate of Table~\ref{tab:params} over the first \num{1700} iterations.

%% file: sections/s4_experiments.tex
\section{Experiments}
\label{sec:results}

\begin{table}[t]
    \caption{Locomotion test of TGI and PPO on the G1, in \% (mean $\pm$ standard deviation over seeds). Bold: best.}
    \label{tab:main}
    \centering\footnotesize
    \setlength{\tabcolsep}{3pt}
    \begin{tabular}{@{}ll cc cc@{}}
        \toprule
        & & \multicolumn{2}{c}{TGI} & \multicolumn{2}{c}{PPO} \\
        \cmidrule(lr){3-4} \cmidrule(l){5-6}
        & & success & falls & success & falls \\
        \midrule
        Newton & Locomotion & $\mathbf{91.8} \pm 1.2$ & $1.0 \pm 0.2$ & $84.9 \pm 2.0$ & $\mathbf{0.6} \pm 0.0$ \\
               & Held out   & $\mathbf{90.9} \pm 1.3$ & $1.5 \pm 0.4$ & $83.4 \pm 2.2$ & $\mathbf{1.0} \pm 0.1$ \\
               & Sparse     & $\mathbf{55.7} \pm 1.4$ & $20.3 \pm 1.3$ & $48.2 \pm 2.6$ & $\mathbf{15.1} \pm 0.9$ \\
        \midrule
        PhysX  & Locomotion & $\mathbf{86.5} \pm 2.2$ & $1.0 \pm 0.2$ & $85.1 \pm 2.3$ & $\mathbf{0.8} \pm 0.1$ \\
               & Held out   & $\mathbf{86.7} \pm 2.6$ & $1.5 \pm 0.2$ & $83.0 \pm 2.2$ & $\mathbf{1.4} \pm 0.3$ \\
               & Sparse     & $\mathbf{54.8} \pm 1.8$ & $18.6 \pm 0.8$ & $48.5 \pm 1.8$ & $\mathbf{15.2} \pm 1.3$ \\
        \bottomrule
    \end{tabular}
\end{table}

\begin{table}[t]
    \caption{Goals reached by ANYmal-D per terrain type, in \% (mean $\pm$ standard deviation over seeds), after training with Newton. Bold: best.}
    \label{tab:anymal}
    \centering\footnotesize
    \setlength{\tabcolsep}{5pt}
    \begin{tabular}{@{}l cc cc@{}}
        \toprule
        & \multicolumn{2}{c}{Newton} & \multicolumn{2}{c}{PhysX} \\
        \cmidrule(lr){2-3} \cmidrule(l){4-5}
        & TGI & PPO & TGI & PPO \\
        \midrule
        All    & $\mathbf{91.0} \pm 2.1$ & $81.2 \pm 1.5$ & $\mathbf{81.6} \pm 2.0$ & $76.4 \pm 1.1$ \\
        Climb  & $\mathbf{79.9} \pm 3.9$ & $73.9 \pm 3.1$ & $\mathbf{66.5} \pm 3.7$ & $64.5 \pm 2.2$ \\
        Stairs & $\mathbf{93.5} \pm 1.6$ & $62.6 \pm 5.2$ & $\mathbf{77.0} \pm 2.0$ & $59.6 \pm 5.3$ \\
        Stones & $\mathbf{77.7} \pm 7.5$ & $70.0 \pm 1.9$ & $\mathbf{69.2} \pm 7.9$ & $67.6 \pm 1.5$ \\
        Pit    & $\mathbf{86.8} \pm 9.8$ & $84.7 \pm 5.5$ & $\mathbf{81.5} \pm 8.7$ & $81.1 \pm 5.0$ \\
        Gap    & $\mathbf{94.4} \pm 1.0$ & $83.5 \pm 3.4$ & $\mathbf{79.7} \pm 1.6$ & $77.9 \pm 2.4$ \\
        Boxes  & $\mathbf{96.1} \pm 1.8$ & $72.1 \pm 1.8$ & $\mathbf{90.6} \pm 3.5$ & $70.0 \pm 1.4$ \\
        \bottomrule
    \end{tabular}
\end{table}

\begin{table}[t]
    \caption{Ablations of TGI on the G1 in the locomotion test with PhysX: success and falls in \% (mean $\pm$ standard deviation over seeds), and 8-step error logged in training (Err.). Bold: best; underlined: second best.}
    \label{tab:ablation}
    \centering\footnotesize
    \setlength{\tabcolsep}{6pt}
    \begin{tabular}{@{}l cc c@{}}
        \toprule
        Variant & Success & Falls & Err. \\
        \midrule
        \multicolumn{4}{@{}l}{\emph{40k iterations}} \\
        TGI & $\mathbf{86.5} \pm 2.2$ & $\mathbf{1.0} \pm 0.2$ & \textbf{0.12} \\
        \quad Model: no $\mathbf{h}$ & $15.8 \pm 1.7$ & $34.5 \pm 1.7$ & 0.19 \\
        \quad No gradient cut & $\underline{86.2} \pm 1.2$ & $2.6 \pm 0.5$ & \textbf{0.12} \\
        \quad Bilinear support & $85.5 \pm 0.8$ & $\underline{1.8} \pm 0.2$ & \textbf{0.12} \\
        \midrule
        \multicolumn{4}{@{}l}{\emph{20k iterations}} \\
        TGI ($H = 32$) & $\mathbf{82.6} \pm 4.0$ & $\mathbf{3.8} \pm 0.9$ & \textbf{0.12} \\
        \quad Horizon $H = 8$ & $56.8 \pm 2.3$ & $31.3 \pm 1.7$ & 0.22 \\
        \quad Horizon $H = 16$ & $78.0 \pm 0.4$ & $\underline{6.9} \pm 1.4$ & \underline{0.13} \\
        \quad Horizon $H = 64$ & $\underline{78.3} \pm 4.7$ & $10.0 \pm 2.0$ & 0.14 \\
        \bottomrule
    \end{tabular}
\end{table}

\begin{table}[t]
    \caption{8-step error with and without $\mathbf{h}$ on rollouts of the trained policies, by terrain and by touchdown within the 8 steps.}
    \label{tab:wm_err}
    \centering\footnotesize
    \setlength{\tabcolsep}{3.5pt}
    \begin{tabular}{@{}l cc @{\hspace{9pt}} l cc@{}}
        \toprule
        Terrain & $\mathbf{h}$ & No $\mathbf{h}$ & Windows & $\mathbf{h}$ & No $\mathbf{h}$ \\
        \midrule
        Flat & \textbf{0.024} & 0.070 & All & \textbf{0.178} & 0.557 \\
        Slopes & \textbf{0.029} & 0.165 & Training types & \textbf{0.165} & 0.510 \\
        Inverted slopes & \textbf{0.029} & 0.153 & Held-out types & \textbf{0.197} & 0.634 \\
        Stairs & \textbf{0.240} & 0.818 & No touchdown & \textbf{0.115} & 0.276 \\
        Inverted stairs & 0.318 & 0.318 & Supported & \textbf{0.146} & 0.554 \\
        Boxes & \textbf{0.192} & 0.677 & Poorly supp. & \textbf{0.824} & 4.009 \\
        Random rough & \textbf{0.322} & 2.515 & & & \\
        Stepping stones & \textbf{0.174} & 0.493 & & & \\
        \bottomrule
    \end{tabular}
\end{table}

We evaluate 2.5D TGI on the G1 with the locomotion test of~\cite{heyrman2026watch} in Isaac Lab, which covers 13 terrain types, five of which are held out from training, in about 2990 episodes.
Sparse terrains include stepping stones, pallets, and concentric gaps.
We test with Newton, which is used for training, and with PhysX to verify sim-to-sim transfer.
An episode is a success if the robot does not fall and covers at least $70\%$ of its commanded distance.
ANYmal-D is evaluated on ten levels of each terrain type by the share of goals that it reaches.

As a baseline, we train policies with PPO in the same simulator and with the same terrains, curriculum, randomization, perception, and reward as the TGI policies.
The two robots test how a training setup carries over between the two methods in opposite directions.
For the G1, the environment, the reward, and the PPO settings are those of~\cite{heyrman2026watch}, which were tuned for PPO, and we transfer them to TGI.
For both methods, we replace the terrain curriculum with success-guided sampling~\cite{zhang2026balanced} and remove the gravity curriculum, which offsets part of the robot's weight at low terrain levels, because a world model struggles to track dynamics that change over training.
For ANYmal-D, the environment and the reward were tuned for TGI and then transferred to PPO with the default PPO settings of Isaac Lab's rough-terrain locomotion task~\cite{nvidia2025isaaclabgpuacceleratedsimulation}.

Both methods train for the same number of iterations, \num{40000} for the G1 and \num{30000} for ANYmal-D, and we report means over 5 training seeds.
Training time is compared for the same number of iterations on the same GPU.
The 8-step error of the world model is the normalized squared error of its prediction after 8 open-loop steps on simulator transitions.

\begin{figure*}[t]
    \centering
    \includegraphics[width=0.95\linewidth]{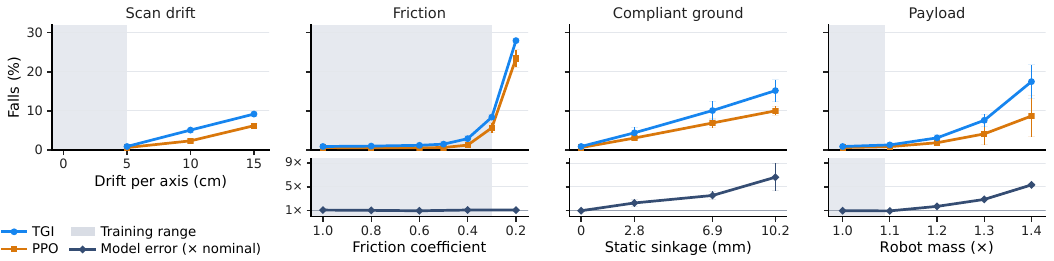}
    \caption{Beyond the training distribution. Top: falls in the locomotion test. Bottom: 8-step error of the world model, relative to the nominal condition. Bars: standard deviation over seeds.}
    \label{fig:ood}
\end{figure*}

\subsection{TGI and PPO}
\label{ssec:res_ppo}

Table~\ref{tab:main} compares TGI and PPO on the G1.
TGI matches the success of PPO with PhysX and exceeds it with Newton, but it falls more often in both, particularly on sparse terrain.
PPO in turn trains more cautious policies, which fall less but cover less of the commanded distance in the episodes without a fall, about $80\%$ against $83\%$ to $86\%$ for TGI.

Fig.~\ref{fig:ood} compares the two methods outside the training distribution, together with the error of the world model as a measure of how far each condition moves the dynamics from those of training.
On the held-out terrain types, TGI keeps its success (Table~\ref{tab:main}).
Under scan drift and low friction, TGI falls about 3 and 5 points more often than PPO, while the error of the model barely changes under low friction.
On compliant ground and at $1.4\times$ the robot's mass, the error grows to $6.6\times$ and $5.3\times$, and TGI falls 5 and 9 points more often.
The two largest gaps to PPO occur where the error of the model grows.

Table~\ref{tab:anymal} shows the goals reached by ANYmal-D policies trained with 3D TGI and the impulse model.
Over the same number of iterations, TGI outperforms PPO at its default settings with both Newton and PhysX, with the largest gains on stairs and boxes.
From Newton to PhysX, TGI loses 9.4 points and PPO 4.8, so its margin is smaller outside the training simulator, as for the G1.

On the same GPU, training the G1 to \num{40000} iterations takes \SI{4.9}{\hour} with TGI and \SI{29.9}{\hour} with PPO, which is $6.1\times$ faster.
Training ANYmal-D to \num{30000} iterations with 3D TGI takes \SI{5.0}{\hour} against \SI{24.4}{\hour}, which is $4.9\times$ faster.
An iteration of TGI also collects $9.6\times$ fewer simulator transitions than one of PPO, 2.5 on average against 24 steps per environment.
On ANYmal-D, whose training was tuned for TGI, TGI reaches the final curriculum level of PPO in about as many iterations and with $10\times$ fewer simulator transitions, 0.23 against 2.36 billion.
On the G1, whose training was tuned for PPO, TGI has similar results, which shows that a training setup tuned for PPO can transfer to training in a world model.

\subsection{Ablations}
\label{ssec:res_ablation}

Table~\ref{tab:ablation} reports ablations over individual TGI settings.
We first remove the terrain features $\mathbf{h}$ from the inputs of the world model~\eqref{eq:model}, while the rewards, the terminations, and the observation still receive them.
The resulting policy succeeds in about one in six locomotion episodes.
Table~\ref{tab:wm_err} reports the error of the two world models.
On rollouts of the trained policies, the error without $\mathbf{h}$ is about $3\times$ higher overall and up to $8\times$ higher on random rough terrain.
The two models match only on inverted stairs, and even on flat terrain the error without $\mathbf{h}$ is about $3\times$ higher, as $\mathbf{h}$ also contains the base height and the foot clearances.
The error without $\mathbf{h}$ is about $5\times$ higher on poorly supported touchdowns.
The robot state alone thus does not determine the motion, and the world model cannot accurately learn environment interactions without terrain information.

Without the gradient cut of~\eqref{eq:gated_return}, the policy has similar success rates but falls in 2.6\% instead of 1.0\% of episodes.
Using bilinear interpolation of the support instead of calculating the nearest node minimally affects the success rate.

We compare imagination horizons after \num{20000} iterations.
We use $H = 32$, which has the highest success rate and the lowest world model error.
A much shorter horizon, $H = 8$, produces a myopic policy that cannot learn from long-horizon success and as a result falls in almost a third of the episodes.
With $H = 16$ the world model performance is similar to $H = 32$, but the policy performance is lower.
At $H = 64$, we attribute the lower performance to a noisier and larger gradient with more contact events and higher compounding model error.

\begin{figure}[t]
    \centering
    \includegraphics[width=\linewidth]{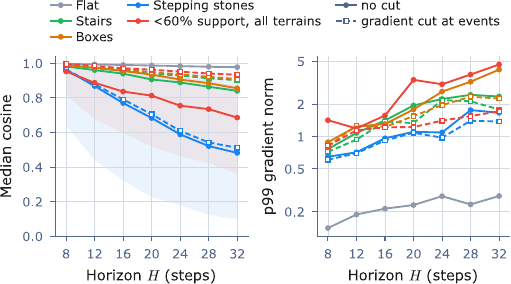}
    \caption{Agreement of the backpropagated gradient with finite differences over the horizon: median cosine (left) and 99th-percentile gradient norm (right). Solid: full return; dashed: return with the gradient cut. Red: starts before touchdowns with less than $60\%$ support.}
    \label{fig:grad}
\end{figure}

\begin{table}[t]
    \caption{Median cosine ($\uparrow$) between the gradient of an objective and three references, at starts before touchdowns with less than $60\%$ support. Bold: best; underlined: second best.}
    \label{tab:controls}
    \centering\footnotesize
    \setlength{\tabcolsep}{4pt}
    \begin{tabular}{@{}l ccc@{}}
        \toprule
        & \multicolumn{2}{c}{Finite difference of} & Smoothed \\
        \cmidrule(lr){2-3}
        Gradient of & its objective & the full return & gradient \\
        \midrule
        Return cut at events (TGI) & \textbf{0.91} & 0.19 & \textbf{0.77} \\
        Full return & 0.65 & \textbf{0.65} & 0.65 \\
        Return cut at random steps & \underline{0.70} & \underline{0.57} & \underline{0.68} \\
        Return over a shorter horizon & 0.67 & 0.41 & 0.66 \\
        \bottomrule
    \end{tabular}
\end{table}

\begin{table}[t]
    \caption{The gradient cut in other settings. F, C: cosine ($\uparrow$) of the full and the cut gradient with the finite difference of their own objective; C/F: cut gradient against the full return. Norm ($\downarrow$): 99th percentile.}
    \label{tab:diag}
    \centering\footnotesize
    \setlength{\tabcolsep}{3pt}
    \begin{tabular}{@{}l c cc ccc cc@{}}
        \toprule
        & Rollouts & \multicolumn{2}{c}{Before touchd.} & \multicolumn{3}{c}{All starts} & \multicolumn{2}{c}{Norm} \\
        \cmidrule(lr){3-4} \cmidrule(lr){5-7} \cmidrule(l){8-9}
        Setting & cut (\%) & F & C & F & C & C/F & F & C \\
        \midrule
        Trained with cut & 12.1 & 0.65 & 0.91 & 0.89 & 0.91 & 0.87 & 2.1 & 1.6 \\
        Trained without cut & 13.1 & 0.66 & 0.92 & 0.90 & 0.92 & 0.89 & 2.0 & 1.6 \\
        \midrule
        $\eta = 0.1$ & 38.8 & 0.68 & 0.96 & 0.90 & 0.96 & 0.82 & 2.4 & 1.1 \\
        $\eta = 0.2$ & 25.6 & 0.67 & 0.93 & 0.90 & 0.94 & 0.87 & 2.1 & 1.4 \\
        $\eta = 0.3$ (TGI) & 13.1 & 0.67 & 0.93 & 0.90 & 0.92 & 0.89 & 2.2 & 1.7 \\
        $\eta = 0.5$ & 0.8 & 0.70 & 0.73 & 0.90 & 0.90 & 0.90 & 2.2 & 2.2 \\
        $\delta = \SI{2}{\centi\metre}$ & 12.8 & 0.71 & 0.94 & 0.90 & 0.92 & 0.89 & 2.9 & 2.5 \\
        $\delta = \SI{8}{\centi\metre}$ & 14.5 & 0.68 & 0.93 & 0.90 & 0.92 & 0.88 & 1.7 & 1.3 \\
        \bottomrule
    \end{tabular}
\end{table}

\subsection{Gradient Analysis}
\label{ssec:res_grad}

The gradient cut assumes that the analytic gradient is unreliable where a foot lands on an edge.
We test this by comparing the backpropagated gradient of the imagined return with respect to the first action against the change of the same imagined return when the action is moved by a small step.
The change is a finite difference that moves each action dimension in turn by $\pm 0.05$, at start states collected with the trained policy on every training terrain.
The cosine of the two gradients is close to 1 if the gradient matches the change and close to 0 if it does not, so a higher cosine indicates a more reliable gradient (Fig.~\ref{fig:grad}).

We separately report starts whose rollout contains an imagined touchdown with less than $60\%$ support, a stricter threshold than that of the cut.
For the full return over 32 steps, the median cosine stays at 0.98 on flat ground but falls below 0.7 at these starts and to 0.48 on stepping stones.
With a $10\times$ smaller step, it stays above 0.80 on every terrain, so the gradient holds for small steps, but the return changes too abruptly at these touchdowns for it to hold over the larger step.
Poorly supported touchdowns are also where the world model is least accurate, with about $6\times$ the error of supported touchdowns (Table~\ref{tab:wm_err}).
Fig.~\ref{fig:grad}, Table~\ref{tab:controls}, and the lower rows of Table~\ref{tab:diag} use a policy after \num{20000} iterations, each on its own sample of start states, and the upper rows of Table~\ref{tab:diag} the policies after \num{40000}.

Table~\ref{tab:controls} shows what the cut changes at these starts.
The smoothed gradient is a randomized-smoothing estimate of the gradient of the same objective.
The gradient of the cut return predicts the change of its own objective with a cosine of 0.91, against 0.65 for the full return.
Cutting at random steps at the same overall rate, or shortening every rollout to the mean cut horizon of 30 steps, does not have this effect, which suggests that the gain comes from where the cut is placed.
The price is that the cut return is a different objective: at these starts, its gradient agrees with the finite difference of the full return at only 0.19.
Over all starts, the two stay close, at 0.87, because only about one rollout in eight is cut (Table~\ref{tab:diag}).
The same holds for a policy trained without the cut.
A lower threshold $\eta$ cuts more rollouts and moves the gradient further from that of the full return.
A finer height map raises the gradient norm.

\subsection{Extensions}
\label{ssec:res_variations}

Since rewards, terminations, and observations are computed from the terrain query and from the predicted states and forces, TGI can be extended with new training signals and terrain styles.
We test two such extensions.

The contact forces of the impulse model~\eqref{eq:force} allow rewards and penalties on forces in imagination, which we use for the impact penalty of ANYmal-D.
The model predicts $99\%$ of the simulated foot contacts, and its foot forces correlate with the simulated ones at $0.90$.
However, rare contacts are predicted less reliably.
The model predicts $81\%$ of the shank contacts, which occur in about $1.5\%$ of the steps.
Impacts are also underestimated.
Forces above \SI{450}{\newton} (the penalty starts at \SI{600}{\newton}) are predicted at a median of $0.65\times$ their simulated size, so a penalty on imagined forces is weaker than the same penalty in the simulator, which may account for part of the margin of TGI over PPO on ANYmal-D.

The terrain query also extends to other perception modalities.
As a demonstration rather than a comparison, we train the G1 with voxel perception based on~\cite{ben2025gallant} and the goal command of ANYmal-D on rough terrain, stepping stones, gaps, low bars, low ceilings, and ledges, in this case with PhysX.
The world model and the optimization follow the shared settings of Table~\ref{tab:params}.
The policy reaches the goal in $99.7\%$ of the test episodes and falls in $0.15\%$.
On the three overhang terrains, it reaches the goal in at least $98.7\%$ of the episodes and ducks under every obstacle.
It touches an obstacle in $0.6\%$ of these episodes.
TGI trains $5.9\times$ more iterations per second than PPO on the same GPU; we do not compare their policies.

\subsection{Hardware}
\begin{figure}[t]
    \centering
    \includegraphics[width=\linewidth]{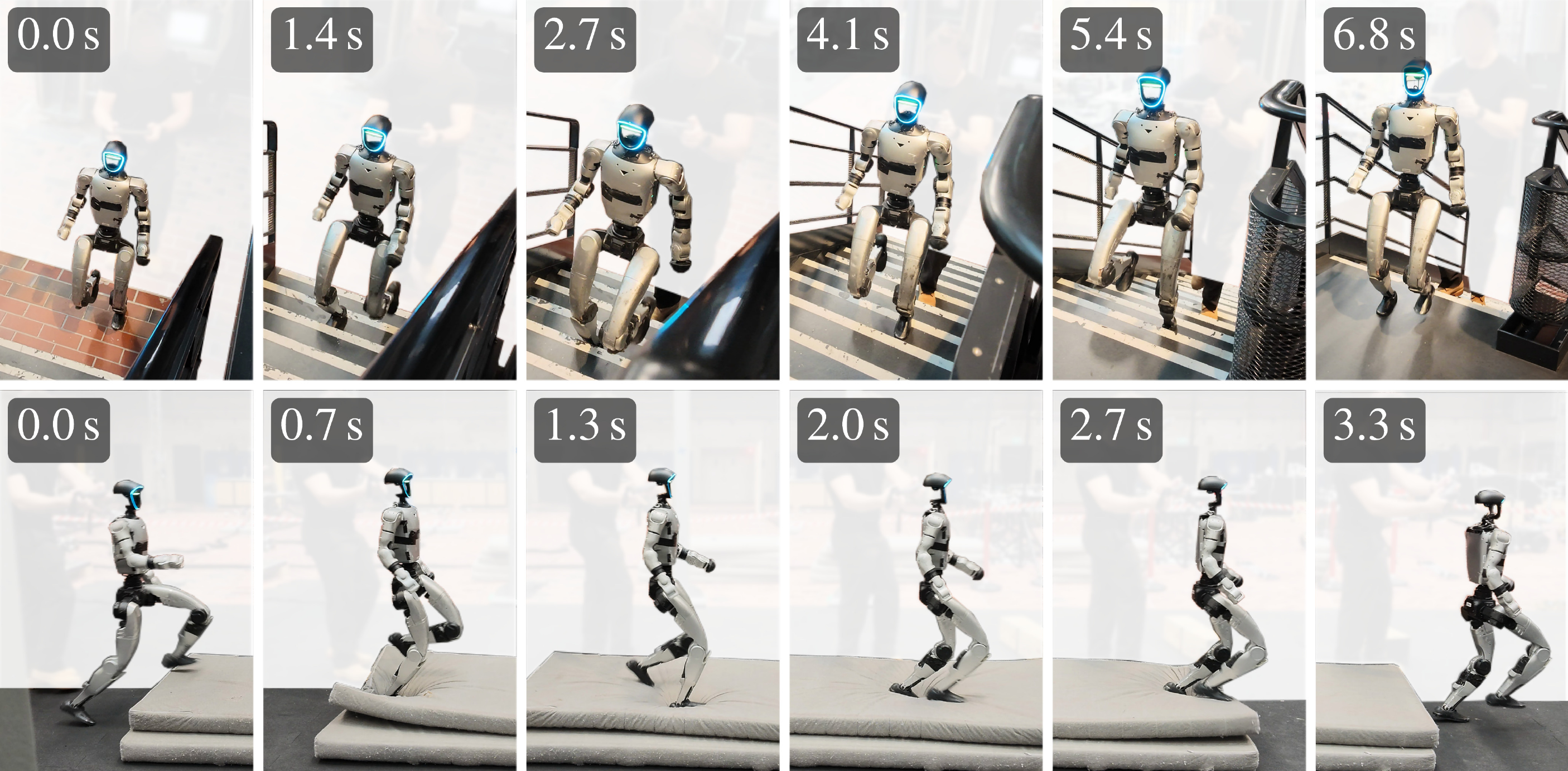}
    \caption{The TGI policy on the G1, on stairs (top) and on two stacked mattresses (bottom). The operator holds a safety strap without pulling on it.}
    \label{fig:hardware}
\end{figure}
The TGI-trained G1 locomotion policy transfers to hardware: it climbs stairs and crosses two stacked mattresses (Fig.~\ref{fig:hardware}), although compliant ground raises its fall rate in simulation (Fig.~\ref{fig:ood}).
As in training, the deployed policy observes height-map patches around the footholds predicted by the FPM~\cite{heyrman2026watch}.
Videos are available on the project page.

%% file: sections/s5_conclusion.tex
\section{Limitations and Conclusion}
\label{sec:conclusion}

We presented terrain-grounded imagination (TGI), a method for training perceptive legged locomotion entirely in a learned world model.
The world model learns only how the robot responds to the terrain, while its terrain features, the rewards, the terminations, and the robot's perception are computed from differentiable queries of the known terrain geometry, so policies with new kinds of terrain and perception are trained by changing the query.

We demonstrated that a humanoid training setup tuned for PPO carries over to TGI, whose policy matches the success of PPO in less training time and transfers to hardware.
For contact-rich quadruped locomotion, a column field and an impulse world model that predicts full-body contact forces let TGI reach more goals than PPO at default settings in less training time.
Cutting the gradient at poorly supported touchdowns reduces the falls of the humanoid, and a demonstration with voxel perception trains a G1 policy that passes under overhangs.
On hardware, the policy crosses stacked mattresses, a deformable terrain unseen in training and not representable by the world model.

However, the TGI world model has several limitations compared to training in simulation.
On the G1 task, which was tuned for PPO, the TGI policy falls more often than the PPO policy.
The gap grows on compliant ground and under heavy payloads, where the error of the world model grows.
As the TGI world model needs to learn the transition function, TGI cannot use curricula that change the dynamics, such as the gravity curriculum of~\cite{heyrman2026watch}.
The predicted contact forces underestimate impacts and are less reliable for rare contacts, which weakens penalties on imagined forces.
Finally, the PPO baselines use existing settings, and we have not compared TGI with off-policy model-free methods or with gradients from a differentiable simulator.
Replacing the known terrain with the robot's own elevation map would allow the world model to be refined on real-world data, which we leave to future work.